\documentclass[10pt, a4paper]{article}

\usepackage[final]{lrec-coling2024} 

\title{CrossTune: Black-Box Few-Shot Classification with Label Enhancement}

\name{Danqing Luo\textsuperscript{\rm 1${\star}$}\thanks{${\star}$Equal contribution.}, Chen Zhang\textsuperscript{\rm 1${\star}$}, Yan Zhang\textsuperscript{\rm 1}, Haizhou Li\textsuperscript{\rm 1,2${\dag}$}\thanks{${\dag}$Corresponding author.}} 

\address{\textsuperscript{\rm 1} National University of Singapore \\ \textsuperscript{\rm 2} The Chinese University of Hong Kong (Shenzhen), China \\
         chen\_zhang@u.nus.edu, \{danqing, eleyanz, haizhou.li\}@nus.edu.sg\\}

\abstract{
Training or finetuning large-scale language models (LLMs) requires substantial computation resources, motivating recent efforts to explore parameter-efficient adaptation to downstream tasks. One approach is to treat these models as black boxes and use forward passes (Inference APIs) to interact with them. Current research focuses on adapting these black-box models to downstream tasks using gradient-free prompt optimization, but this often involves an expensive process of searching task-specific prompts. Therefore, we are motivated to study black-box language model adaptation without prompt search. Specifically, we introduce a label-enhanced cross-attention network called CrossTune, which models the semantic relatedness between the input text sequence and task-specific label descriptions. Its effectiveness is examined in the context of few-shot text classification. To improve the generalization of CrossTune, we utilize ChatGPT to generate additional training data through in-context learning. A switch mechanism is implemented to exclude low-quality ChatGPT-generated data. Through extensive experiments on seven benchmark text classification datasets, we demonstrate that our proposed approach outperforms the previous state-of-the-art gradient-free black-box tuning method by 5.7\% on average. Even without using ChatGPT-augmented data, CrossTune performs better or comparably than previous black-box tuning methods, suggesting the effectiveness of our approach.
 \\ \newline  }

\begin{document}

\maketitleabstract

\section{Introduction}

\noindent In the past few years, significant progress has been made in research on large-scale language models (LLMs)~\citep{devlin-etal-2019-bert,liu2019roberta,ouyang2022training,chowdhery2022palm}. Scaling up language models has been demonstrated to boost performance and sample efficiency on a great variety of downstream tasks~\citep[\textit{inter alia}]{raffel-etal-2020-exploring,brown-etal-2020-language}. However, training such LLMs is not practical with typical research hardware. Even finetuning them on task-specific data is extremely challenging. Many research efforts have been devoted to more parameter-efficient adaptation approaches, including (1) parameter-efficient fine-tuning (PEFT)~\citep{lester-etal-2021-power,li-liang-2021-prefix,houlsby-etal-2019-parameter,hu2022lora}, which optimizes a small portion of task-specific parameters, while keeping the language model intact; (2) prompt-based learning, where a carefully-designed task-specific sequence, known as a prompt, is added to the input text sequence of a pre-trained language model (LM). The LM is repurposed to adapt to the downstream tasks without additional training.

Due to commercial reasons, powerful LLMs are provided as a service in the cloud, and end users can only interact with them through inference APIs. This setup is referred to as Language-Model-as-a-Service (LMaaS)~\citet{sun-etal-2022-bbt}. Popular PEFT approaches are impractical in this context since they require access to model gradients. To address this challenge, an emerging line of prompt-based learning research focuses on gradient-free prompt optimization techniques~\citep{brown-etal-2020-language,sun-etal-2022-bbt,sun-etal-2022-bbtv2,deng-etal-2022-rlprompt,prasad-etal-2023-grips,pmlr-v202-hou23b}. However, these methods are also problematic because (1) prompt optimization is highly sensitive to the template design and demonstration selection~\citep{gao-etal-2021-making,zhao-etal-2021-calibrate} leading to unstable performance and poor generalization. (2) The prompt search process, either manual or automatic, is also time-consuming. For example, the covariance matrix adaptation evolution strategy (CMA-ES) adopted by~\citet{sun-etal-2022-bbt} requires tens of thousands of forward passes through the LLMs to achieve satisfactory performance even in few-shot text classification scenarios. 

To this end, we propose CrossTune, a label-enhanced black-box few-shot learner for the adaptation of the black-box LMs without prompt search. Following existing works, we assume the inference APIs provide forward-pass LM outputs and study our approach in the context of few-shot text classification. In CrossTune, the black-box model is treated as a feature extractor where hidden states of the input text sequence are derived. Besides, the original label words are expanded to long text descriptions. A cross-attention network is trained to align the input text sequence with its associated label. In this way, we can steer the model to focus on specific aspects of the input text data that are semantically related to the label descriptions, which act as a form of contextual input and provide additional semantic guidance to the model about what each label means. 

In the few-shot scenarios, the model can easily overfit the training data resulting in poor generalization to unseen test data. Existing works mainly rely on semi-supervised and weakly-supervised methods to boost the generalization of the text classifiers. Both assume the presence of abundant in-distribution unlabeled data~\citep{schick-schutze-2021-exploiting,chen-etal-2021-revisiting,fei-etal-2022-beyond,du-etal-2021-self,cho-etal-2023-celda}\footnote{The unlabeled data are either the original training set with their ground-truth labels removed or retrieved sentences from a sentence bank based on their similarity to the few-shot training examples.}. Contrary to prior works, we do not make such an assumption. Instead, we harness the strong instruction-following capability of ChatGPT\footnote{\url{https://openai.com/chatgpt}} to generate data conditioned on the labels through in-context learning~\citep{brown-etal-2020-language}. To filter out low-quality generation, we implement an additional switch mechanism as described in \S\ref{subsec:swich-mechansim}. 


\bigskip
\noindent
In summary, our contributions are as follows: 

\begin{itemize}
    \item We introduce CrossTune, a new approach for the few-shot adaptation of black-box language models. Different from existing methods, CrossTune does not rely on the expensive prompt search process. Additionally, CrossTune leverages the rich semantic information in label descriptions to perform the classification task.
    
    \item Instead of relying on in-distribution unlabeled training data, which are rarely available in real-life scenarios, we harness the power of a strong instruction-following text generator, ChatGPT, to generate data conditioned on the labels through in-context learning. A pipeline is designed to generate and clean the data. Our experiments demonstrate that the quality of data generated by ChatGPT is on par with the original training data. 

    \item Extensive experiments are performed on 7 few-shot text classification datasets and CrossTune significantly outperforms previous the state-of-the-art gradient-free prompt optimization approach with an absolute improvement of 5.7\% on average.
\end{itemize}

\section{Related Work}

\paragraph{Gradient-Free Black-Box Tuning}
The success of prompt-based learning with GPT-3~\citep{brown-etal-2020-language} has inspired fruitful research in NLP community.  A typical line is to optimize the prompts for downstream tasks based on the gradients of pretrained language models such that the output can align closely with the desired results~\citep{gao-etal-2021-making,chen-etal-2021-revisiting,li-liang-2021-prefix,liu2021gpt}.  However, many practical applications involve models where internal parameters or gradients are obscured or inaccessible, leading to a so-called "black-box" tuning setting~\citep{sun-etal-2022-bbt,diao2023blackbox}. 

Several studies have ventured into black-box tuning challenges. BBT~\citep{sun-etal-2022-bbt} and BBTv2~\citep{sun-etal-2022-bbtv2} utilize the CMA evolution strategy to optimize prompts but face challenges in efficiency and flexibility. RLPrompt~\citep{deng-etal-2022-rlprompt} and Black-box Discrete Prompt Learning (BDPL)~\citep{diao2023blackbox} both use reinforcement learning to fine-tune discrete prompts, with BDPL featuring a streamlined search approach. TEMPERA~\citep{zhang2023tempera} expands optimization components, while GrIPS~\citep{prasad-etal-2023-grips} focuses on phrase-level editing. However, many of these black-box tuning methods suffer from efficiency issues and may not always deliver optimal results. Recently, PromptBoosting ~\citep{pmlr-v202-hou23b} adapts the ensembling idea of AdaBoost to black-box tuning and achieves state-of-art performance in multiple black-box few-shot classification tasks. Different from the existing approaches, CrossTune does not require the expensive prompt search and offers a much simpler and more effective adaption of the black-box language models. 

\paragraph{Few-shot Text Classification with Augmented Data} 

Popular research directions for enhancing the generalization of few-shot text classifier include semi-supervised learning~\citep{xie-etal-2020-uda,sohn2020fixmatch,zoph2020rethinking} and weakly-supervised learning~\citep{meng-etal-2020-text,zhang-etal-2021-weakly,fei-etal-2022-beyond,cho-etal-2023-celda}. Both line of works assume the presence of a substantial amount of unlabeled data. Common techniques to obtain unlabeled text data include (1) removing the gold labels of the original full training data for a specific task~\citep{chen-etal-2021-revisiting,schick-schutze-2021-exploiting}, (2) applying a retriever to retrieve sentences from a large-scale sentence bank that are semantically similar to the few-shot training data~\citep{du-etal-2021-self}, and (3) text data augmentation, such as paraphrasing and back-translation~\citep{10.1145/3544558}. However, these techniques have several limitations: Using the full training data as an unlabeled source is often impractical because substantial in-distribution unlabeled data is not always available in real-life scenarios. Moreover, retrieval and text augmentation tend to produce similar unlabeled data to the few-shot training set, limiting the diversity of the augmented data. Furthermore, both semi- and weakly-supervised learning rely on potentially inaccurate pseudo-labeling of the unlabeled data. 

Motivated by the recent imitation learning research on distilling high-quality training data from strong LLMs, like ChatGPT and GPT-4~\citep{wang-etal-2023-self-instruct,xu2023wizardlm,mukherjee2023orca}, we tackle the above limitations by prompting ChatGPT to generate high-quality training data through in-context learning. With its strong instruction-following and text-generation capabilities, ChatGPT serves as a powerful tool for text data augmentation.

\section{Methodology}
\begin{figure}[!t]
  \centering
  \includegraphics[width=\linewidth]{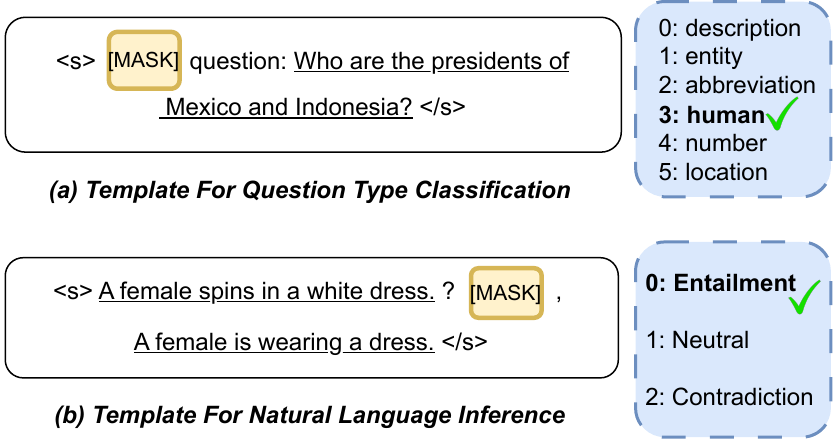}
  \caption{Input template examples. The blue boxes contain the labels for the corresponding classification tasks. }
  \label{fig:input-example}
\end{figure}

\subsection{Problem Formulation}

\noindent In a few-shot text classification task $T$ with a label space $\mathcal{Y}$, we assume there are $K$ labeled training examples per class in the training set, $\mathcal{D}^T_{train}$. The training data size, $|\mathcal{D}^T_{train}| = K \times |\mathcal{Y}|$. We also assume an development set, $\mathcal{D}^T_{dev}$, which is of equal data size as $\mathcal{D}^T_{train}$. Both $\mathcal{D}^T_{train}$ and $\mathcal{D}^T_{dev}$ consist of data instances $(X_i, y_i)$ where $y_i\in{\mathcal{Y}}$ and $X_i$ denotes the input text sequence, which contains $n$ tokens, i.e., $X_i = \{x_i^1, x_i^2, \ldots, x_i^n\}$. Assume that we have task-specific template mapping function $\mathcal{F}_T$, which maps $X_i$ to a specific input format $\mathcal{F}_T(X_i)$. 
Figure~\ref{fig:input-example} shows two examples of $\mathcal{F}_T(X_i)$.
The underlined texts in the boxes are the original input texts, $X_i$. Note that no additional prompt token is prepended to the transformed input. Moreover, assume a black-box language model denoted as $\mathcal{M}$, which is for inference only. Through its API, we can obtain the logits of ``[MASK]" tokens and the hidden states of the input text sequences. 
Our goal is to develop a model that generalizes well to an unseen test set $\mathcal{D}^T_{test}$. 

\subsection{CrossTune Architecture}

\begin{figure*}[!t]
    \centering
    \includegraphics[width=0.9\linewidth]{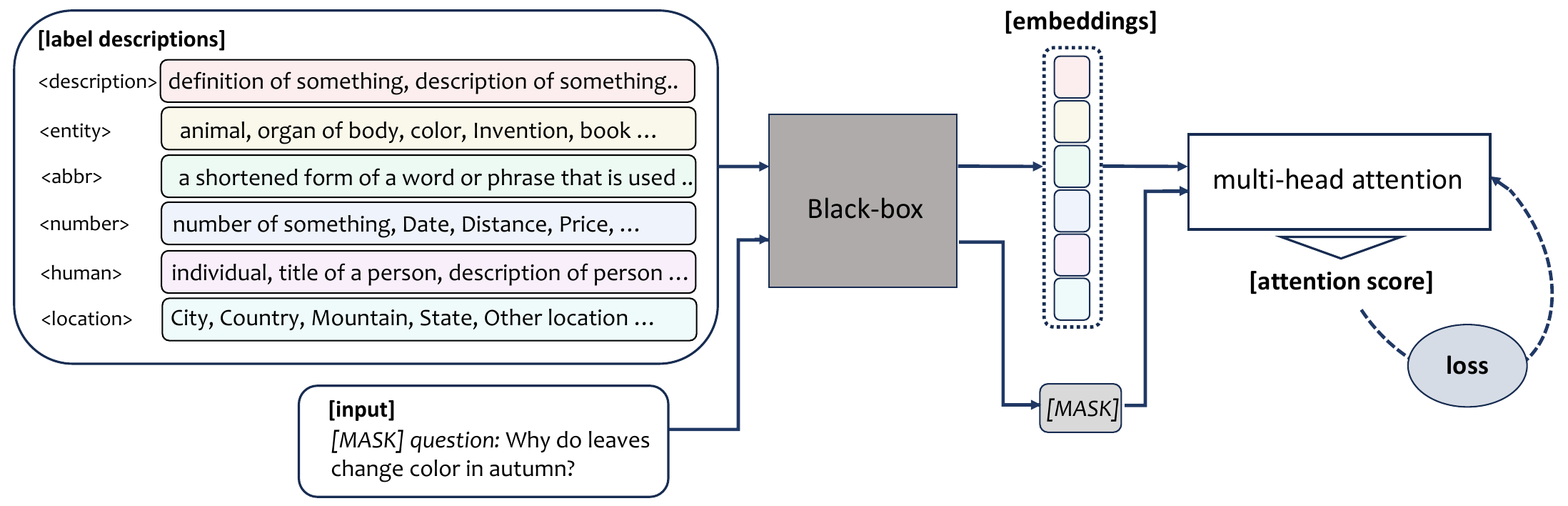}
    \caption{System Overview of CrossTune.} 
    \label{fig:system-overview}
\end{figure*}

\noindent  Figure~\ref{fig:system-overview} presents the model architecture of CrossTune. Using the frozen black-box language model \( \mathcal{M} \), we derive a sequence of hidden states after each layer $l$ with respect to the reformatted input text \( \mathcal{F}_T(X_i) \). As we are interested in the hidden vectors of the ``[MASK]" token that is $\{\textbf{h}_{i,l}^{mask} \in \mathbb{R}^d \}_{l=1}^{L}$, we perform max pooling on $\{\textbf{h}_{i,l}^{mask}\}_{l=L-3}^L$ to derive a single vector representation, $\textbf{h}_{i}^{mask} \in \mathbb{R}^d$. \textcolor{black}{This operation is motivated by previous works on sentence representation learning~\citep{ethayarajh-2019-contextual,li-etal-2020-sentence,hosseini-etal-2023-bert} which state that combining embeddings from multiple layers leads to better semantic representation than using only the last-layer embedding.}

Furthermore, each label in the space \( \mathcal{Y} \) is converted into its corresponding label text description, which is either the definitions specified in the original datasets or from Wikipedia. Using the same model \( \mathcal{M} \), we obtain the single-vector label embeddings \( \mathbf{h}_{y_i} \) for each \( y_i \) in \( \mathcal{Y} \). The \( \mathbf{h}_{y_i} \) embeddings are obtained by applying the same max pooling procedure described above on the hidden states of the label description and then followed by a token-level mean pooling operation.

A multi-head cross-attention module is implemented such that $\mathbf{h}^{mask}_{X_i}$ can attend to each $\mathbf{h}_{y_i}$ in $\mathcal{Y}$. More specifically, $\mathbf{h}^{mask}_{X_i}$ and all label embeddings, $\mathbf{H}_{\mathcal{Y}}$, are first linearly transformed into query vector and key matrix for each head:
   \[ \mathbf{q}^k = W^k_Q\mathbf{h}^{mask}_{X_i} \]
   \[ \mathbf{K}^k = W^k_K\mathbf{H}_{\mathcal{Y}} \]
where \( k \) denotes the k-th head. \( W^k_Q\in{\mathbb{R}^{d \times d}}, W^k_K\in{\mathbb{R}^{d \times d}}\) are the k-th head weight matrices for the query and key respectively. The cross attention is then computed as:

\[ \text{CrossAttn}^k(\mathbf{q}^k, \mathbf{K}^k) = \text{softmax}\left(\frac{\mathbf{q}^k (\mathbf{K}^k)^T}{\sqrt{d}}\right) \]
where d is the dimensionality of the weight matrices. To obtain the final attention scores, we average the scores from each head. These resulting attention scores indicate the significance of each label description in relation to the input text sequence. Finally, cross entropy loss is chosen as the training objective:

\[ \mathcal{L} = \sum_{(X_i, y_i)\in{\mathcal{D}^T_{train}}}{-y_i\ log\  {\hat{y}_i}} \]
where $y_i$ is converted to one-hot vector while $\hat{y}_i$ is the final attention score vector, i.e., probability distribution across the labels in the label space $\mathcal{Y}$. 

\begin{table*}[!ht]
\centering
\resizebox{\linewidth}{!}{

\begin{subtable}[c]{0.55\textwidth}
\centering
    \colorbox{gray!8}{
    \begin{tabular}{@{}p{7.3cm}}
    \#\#\# Instruction:\\
    \{Label\} is defined as \{Label Definition\}. \\\\
    Follow the below examples and generate 10 diverse questions of \{Label\} type and output one question at a line. \\\\

    \#\#\# Examples:\\
    \{Here are the in-context exemplars\} \\\\

    \#\#\# Your Output:\\
    \{Here are the ChatGPT-generated texts\}
    \\\\
    \end{tabular}}
\subcaption{Question Type Classification}
\end{subtable}

\begin{subtable}[c]{0.55\textwidth}
\centering
    \colorbox{gray!8}{
    \begin{tabular}{@{}p{7.3cm}}
    \#\#\# Instruction:\\
    \{Label\} is defined as \{Label Definition\}. \\\\
    Generate 10 diverse \{Label\} sentence pairs in the following format: $[$Premise $|$ Hypothesis$]$ and output one pair at a line.\\\\

    \#\#\# Examples:\\
    \{Here are the in-context exemplars\} \\\\

    \#\#\# Your Output:\\
    \{ChatGPT Generated Text\}
    \\\\
    \end{tabular}}
\subcaption{Natural Language Inference}
\end{subtable}

}\caption{Example instruction templates for prompting ChatGPT to generate task-specific data conditioned on a specific label. In ``{Label Definition}", we provide the meaning of the label. For instance, ``Entailment is defined as when the hypothesis can be logically inferred or implied from the premise" in the case of natural language inference.\label{tab:llm-ins-template}}
\end{table*}

\subsection{ChatGPT for Data Generation}

\noindent We propose to generate task-specific data with ChatGPT (gpt-3.5-turbo). Task-specific instruction templates are designed to prompt ChatGPT to generate relevant text data belonging to a specific class. ChatGPT offers richer data variations, and through in-context learning, it can be prompted for task- and context-specific text generation, ensuring more precise and natural outputs. Table~\ref{tab:llm-ins-template} presents two examples of how we prompt ChatGPT to generate training data. The in-context exemplars are the task- and seed-specific few-shot training data associated with a particular class for which we aim to perform data augmentation with ChatGPT. \textcolor{black}{For selecting the in-context exemplars, we follow the most common setup, which is random sampling, i.e., to generate samples of a particular class, we random sample 8 training samples of that class.} When calling the inference API of ChatGPT, we set the temperature, top\_p, frequency\_penalty, and presence\_penalty to 0.8, 0.95, 0.8, and 1.4 respectively. For each class, we iteratively call the inference API until a sufficient amount of training data is obtained.

\subsection{The Switch Mechanism}
\label{subsec:swich-mechansim}

\noindent Even though ChatGPT is a strong instruction-following text generator, it does not always guarantee the production of high-quality labeled data. Therefore, we utilize the text-understanding capability of another teacher model. \textcolor{black}{We select DeBERTa-base as the teacher model due to its manageable size (suitable for a standard research GPU) and its superior performance on popular text classification benchmarks~\cite{wang2018glue} compared to similar sized models, such as BERT~\citep{devlin-etal-2019-bert}, RoBERTa~\citep{liu2019roberta}, and ELECTRA~\citep{clark2020electra}.} A switch mechanism is introduced such that both ChatGPT and DeBERTa teachers can complement each other and collaboratively determine the labels of the augmented data. Let $\mathcal{A}_{chagpt}$ and  $\mathcal{A}_{deberta}$ denote the ChatGPT and DeBERTa teachers respectively. Motivated by the findings in previous works~\cite{gao-etal-2021-making,chen-etal-2021-revisiting} that prompt-based finetuning of the language model with demonstrations can drastically outperform standard fine-tuning procedures in the low resource setting, we apply prompt-based finetuning for adapting $\mathcal{A}_{deberta}$ to task $T$. The parameter size of \( \mathcal{A}_{deberta} \) is significantly smaller than black-box LM, thus it can be viewed as a readily accessible auxiliary model designed to enhance the quality of the augmented data. Our experimental results reveal that incorporating a switch mechanism with \( \mathcal{A}_{deberta} \) enhances the performance of CrossTune.

\begin{figure}[!ht]
  \centering
  \includegraphics[width=0.9\linewidth]{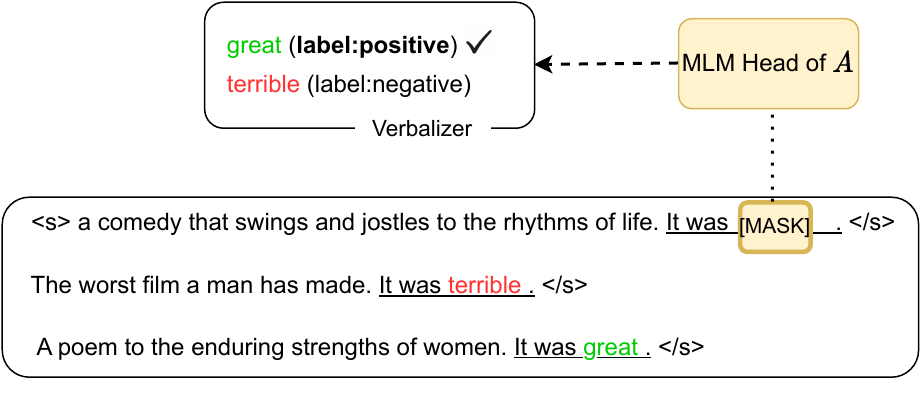}
  \caption{Prompt-based finetuning of \( \mathcal{A}_{deberta} \). The underlined text is the prompt template. In the bottom box, the first, second, and third lines are the input text sequence, the demonstration for label:negative, and the demonstration for label:positive respectively. The verbalizer maps the labels to the corresponding words.}
  \label{fig:prompt-based-learning}
\end{figure}

\paragraph{Prompt-based Finetuning of \( \mathcal{A}_{deberta} \)} 
Figure~\ref{fig:prompt-based-learning} illustrates how we finetune \( \mathcal{A}_{deberta} \). Given $(X_i, y_i) \in \mathcal{D}^T_{train}$, the $X_i$ is first transformed into $\mathcal{F}_T(X_i)$ according to the task-specific templates\footnote{In our experiments, we use the same set of task-specific manual templates for both prompt-based finetuning of $\mathcal{A}_{deberta}$ and the training of CrossTune.}. The verbalizer converts $y_i$ to the corresponding word in the vocabulary of $\mathcal{A}_{deberta}$. To fill in the ``[MASK]" position in $\mathcal{F}_T(X_i)$, $\mathcal{A}_{deberta}$ learns to assign a higher probability to the word mapped to $y_i$ than other label words. For example, $\mathcal{A}_{deberta}$ should predict a higher probability of ``great" than ``terrible" for the example input in Figure~\ref{fig:prompt-based-learning}. To further enhance the prompt-based finetuning process, we append demonstrations after $\mathcal{F}_T(X_i)$. A demonstration is an input text example. For each category, one demonstration is added. $\mathcal{A}_{deberta}$ is finetuned with the standard MLM loss on $\mathcal{D}^T_{train}$. In addition, for model selection, we perform the grid search procedure on different training hyperparameters. The checkpoint with the best performance on $\mathcal{D}^T_{dev}$ is used as the teacher model.

\paragraph{Switch Rule} The data generated by \( \mathcal{A}_{chagpt} \) is equipped with pseudo labels that it deems correct. To validate these labels, we implement a rule to decide if \( \mathcal{A}_{deberta} \) should annotate the data. The decision is based on the classification performance of both \( \mathcal{A}_{chagpt} \) and \( \mathcal{A}_{deberta} \) on \( \mathcal{D}^T_{dev} \). If \( \mathcal{A}_{chagpt} \) outperforms \( \mathcal{A}_{deberta} \), we retain the pseudo labels. Otherwise, we employ \( \mathcal{A}_{deberta} \) for further annotations, discarding any data on which \( \mathcal{A}_{deberta} \) and \( \mathcal{A}_{chagpt} \) disagree and keeping those \(\mathcal{A}_{deberta} \) is confident about.

\section{Experiment Setup}

\paragraph{Datasets} CrossTune is evaluated on 7 text classification datasets, including 3 single-sentence and 4 sentence-pair classification datasets. Among them, AGNews~\cite{zhang-etal-2015-character} is for topic classification. SST-2~\cite{wang2018glue} is for sentiment analysis. TREC~\cite{hovy-etal-2001-toward} is for question classification. MRPC~\cite{wang2018glue} and QQP~\cite{wang2018glue} are paraphrasing tasks. QNLI~\cite{wang2018glue} and MNLI~\cite{bowman-etal-2015-large} are for natural language inference. Following~\citet{sun-etal-2022-bbtv2}, K samples are randomly drawn from the original training set for each class to construct the training set and another K samples from the original training set for the development set. For the test sets, we use the original development set if it exists, otherwise, the original test set is used. K is set to be 16 across all datasets.  

\paragraph{Baselines} We compare our approach with full-model fine-tuning methods and state-of-the-art black-box tuning methods described as follows: (1) \textbf{Finetuning}, the standard way of finetuning a language model for few-shot text classification. (2) prompt-based fine-tuning as implemented by Gao et al.(2021). The approach is referred to as \textbf{LM-BFF}. Both (1) and (2) require updating the weights of the LLM. Hence, they can be seen as white-box methods. (3) \textbf{ICL-RoBERTa}, which applies the in-context learning approach proposed in Brown et al. (2020)~\cite{brown-etal-2020-language} with RoBERTa-large. (4) \textbf{Black-Box Tuning (BBT)}~\cite{sun-etal-2022-bbt}. (5) \textbf{BBTv2}~\cite{sun-etal-2022-bbtv2}. (4) and (5) are derivative-free optimization methods that are based on the covariance matrix adaptation evolution strategy to optimize the continuous prompt~\cite{hansen-ostermeier-2001-completely}. (6) \textbf{RLPrompt}~\citep{deng-etal-2022-rlprompt}, which optimizes the discrete prompts with reinforcement learning and adopts Q-learning to find the best prompt. (7) \textbf{Promptboosting}~\citep{pmlr-v202-hou23b}, which searches the verbalizer and ensemble hundreds of verbalizers via AdaBoost to weight different training samples. (8) To validate the effectiveness of CrossTune, we consider another feature-based variant, which is implemented as an MLP classifier. Specifically, the MASK token embedding is extracted from the frozen black-box model and fed to a 2-layer MLP for classification. We name the baseline MLP-Classifier.

\paragraph{Implementation Details} To align with previous studies on black-box tuning, we employ RoBERTa-Large~\cite{liu2019roberta} (with 354 million parameters) as our large-scale black-box language model. It is important to note that our methodology is model-agnostic. This means that the black-box LLMs can be any encoder-only or encoder-decoder models, even those with billions of parameters. 


\begin{table}[!h]
\centering
\resizebox{0.8\linewidth}{!}{
    \begin{tabular}{l|ll}
    \toprule
    Task Name & Template  \\
    \midrule
    TREC & [MASK] question: $<$X$>$  \\
    AGNews & [MASK] News: $<$X$>$ \\
    SST-2 & $<$X$>$ . It was [MASK] .  \\
    MRPC & $<$$X_1$$>$ ? [MASK] , $<$$X_2$$>$  \\
    QQP & $<$$X_1$$>$ ? [MASK] , $<$$X_2$$>$  \\
    QNLI & $<$$X_1$$>$ ? [MASK] , $<$$X_2$$>$  \\
    MNLI & $<$$X_1$$>$ ? [MASK] , $<$$X_2$$>$ \\
    \bottomrule
    \end{tabular}
}\caption{Task-specific prompt templates and label words.\label{tab:template}}
\end{table}


For training the teacher model $\mathcal{A}_{deberta}$, we set the training batch size, the maximum sequence length, and the maximum number of training steps as 2, 128, and 2000 respectively. We perform grid search on the learning rate of (1e-5, 2e-5) and gradient accumulation steps (1, 2) respectively. \textcolor{black}{The DeBERTa finetuning is conducted on a Nvidia 1080 card, utilizing 5GB of GPU memory. The time cost is quite light. The average hyperparameter search time for a seed is about 30 minutes.} When filtering the ChatGPT-augmented data with the DeBERTa teacher, \textcolor{black}{we set the confidence threshold of the output probability to 0.9 according to its distribution,} preserving up to M samples for each class. Empirically, $1000<=M<=1500$. Table~\ref{tab:data amount} presents the statistics of the data used in our experiment. To ensure a fair comparison, the baseline MLP-Classifier model is trained on the same data as CrossTune.


\begin{table}[!t]
\centering
\resizebox{\linewidth}{!}{
    \begin{tabular}{l|lll}
    \toprule
    Task Name & \#classes & \#augmented data & \#filtered data  \\
    \midrule
    TREC & 6 & 8400 & ~5500 \\
    AGNews & 4 & 7000 & ~4540\\
    SST-2 & 2 & 3700 & ~2800  \\
    MRPC & 2 & 4000 & ~2000  \\
    QQP & 2 & 2800 & ~1900  \\
    QNLI & 2 & 3000 & ~2000  \\
    MNLI & 3 & 10000 & ~2500 \\
    \bottomrule
    \end{tabular}
}
    \caption{The amount of augmented data and filtered data. The data quantity in the table represents the total count across all categories.}
    \label{tab:data amount}
\end{table}

For training CrossTune, we set the train batch size, the learning rate, the total number of training epochs, and the maximum sequence length as 32, 4e-5, 100, and 512 respectively. Grid search is performed on the number of attention heads (1, 2, 4, 8). The model is evaluated with the development set at the end of each epoch and if the validation performance does not improve for consecutive 5 epochs, we early stop the training process. Table~\ref{tab:template} describes the templates we use for training CrossTune. It is worth noting that we do not need a verbalizer in our approach and no additional prompt is prepended to the template. In Figure~\ref{fig:system-overview}, we present the label descriptions of TREC and those of the remaining datasets will be included in the Appendix of the final version.

\begin{table*}[!t]
\centering
\resizebox{\linewidth}{!}{
    \begin{tabular}{l|ccccccc|c}
    \toprule
     & TREC & AGNews & SST-2 & MRPC &  QQP & QNLI & MNLI & Average \\
     & acc & acc & acc &  f1 &  f1 & acc & acc &  \\
    \midrule
    Finetuning$\dagger$  & 88.8 (2.1) & 86.2 (1.4) & 81.4 (3.8) &   76.6 (2.5) &  60.7 (4.3) & 56.3 (1.5) &  45.8 (6.4) & 70.8  \\
    LM-BFF$\dagger$ & 83.4 (2.7) &  87.1 (1.2) & 92.3 (1.5) &  77.8 (2.0) &   69.8 (1.8) & 64.4 (4.6) & 68.7 (2.0) &  77.6 \\ \midrule
    ICL-RoBERTa$\ddagger$ & 26.2 (2.4) & 62.2 (13.5) & 85.9 (0.7)  & 45.8 (6.7) & 36.1 (5.2)  & 53.8 (0.4) & 52.0 (0.7) & 51.7  \\
    BBT$\ddagger$ & 39.3 (5.2) & 81.2 (2.7) & 88.2 (1.7) & 61.6 (4.3) &  48.6 (8.3) & 56.8 (2.0) & 42.3 (2.8) & 59.7 \\
    BBTv2$\ddagger$ & 42.0 (4.5) & 85.3 (0.5) & 83.8 (0.8) & 77.0 (4.7) & 56.3 (3.9)  & 66.3 (2.3) & 51.4 (3.3) & 66.0 \\ 
    RLPrompt$\ddagger$ & 37.3 (3.5) & 76.2 (2.7) & \textbf{90.5} (1.2) & 68.9 (2.1) & 53.7 (2.2) & 52.1 (2.9) & 40.7 (4.7) & 59.9 \\
    PromptBoosting$\ddagger$ & \underline{81.6} (4.0) & 85.2 (0.9) & 87.6 (3.0) & 70.5 (2.9) & \underline{64.8} (3.7) &  58.0 (3.3) & 52.5 (1.5) & 71.5 \\ \midrule
    MLP-Classifier$\ddagger$ & 80.8 (0.2) & \underline{85.9} (0.5) & 89.1 (2.3) & \underline{80.4} (0.5) & \underline{64.8} (2.1) & \underline{70.4} (1.3) & \underline{56.7} (1.5) & \underline{75.4}  \\
    CrossTune$\ddagger$ & \textbf{85.0} (1.8) & \textbf{86.6} (1.1) & \underline{90.2} (2.5) & \textbf{82.3} (0.6) & \textbf{66.1} (1.8) & \textbf{71.4} (0.8) & \textbf{58.5} (1.8) & \textbf{77.2}   \\
    \bottomrule
    \end{tabular}
}
    \caption{Main experiment results. $\dagger$ refers to white-box methods while $\ddagger$ refers to black-box methods. In the black-box category, the best score for each task is highlighted in bold and the second best is underlined. }
    \label{tab:exp-results}
\end{table*}

\section{Results \& Analysis}

\subsection{Main Results} 

\noindent Table~\ref{tab:exp-results} summarizes the main results. We can observe that on average, CrossTune outperforms BBTv2 by 9.4\% on average. It also matches the performance of LM-BFF, which is a strong white-box adaptation method employing prompt-based tuning. Compared to the current SoTA black-box tuning approach, PromptBoosting, CrossTune achieves significantly better results on MRPC, QNLI, and MNLI. It outperforms PromptBoosting by an absolute margin of 5.7\% on average. 

Compared to MLP-Classifier, which also does not rely on the expensive prompt search process and is trained on the same augmented data, CrossTune achieves an improvement of 1.8\% on average across the seven datasets, underscoring that our proposed label cross-attention network is more effective than using an MLP classifier. Furthermore, CrossTune is more lightweight and efficient than MLP-Classifier as their numbers of trainable parameters are 2.10M and 3.15M respectively.

Additionally, we can see that the performance of CrossTune is more consistent with a smaller standard deviation across different data splits compared to prompt-based black-box methods, such as BBTv2 and RLPrompt, suggesting that CrossTune is less likely to overfit to specific data splits and exhibits better generalization.

\begin{table*}[!t]
\centering
\resizebox{0.9\linewidth}{!}{
    \begin{tabular}{l|c|ccccccc|c}
    \toprule
    \multirow{2}{*}{\#data} & \multirow{2}{*}{model} & TREC & AGNews & SST-2 & MRPC &  QQP & QNLI & MNLI & Average \\
    & & acc & acc & acc &  f1 &  f1 & acc & acc &  \\
    \midrule
    \multirow{2}{*}{0} & MLP-Classifier& 46.1 & 82.3 & 88.9 & 76.2 & 64.8 & 54.0 & 53.6 & 66.6  \\
     & CrossTune& 46.4 & 82.5 & 88.2 & 79.5 & 63.8 & 58.8 & 52.5  & 67.4  \\
    \midrule
    \multirow{2}{*}{300} & MLP-Classifier & 73.9 & 85.8 & 88.9 & 78.9 & 67.0 & 67.3 & 54.2 & 73.7  \\
     & CrossTune & 78.6 & 85.1 & 88.6 & 81.8 & 65.7 & 69.1 & 54.7 &  74.8  \\
    \midrule
    \multirow{2}{*}{full} & MLP-Classifier & 80.8  & 85.9  & 89.1  & 80.4 & 64.8 & 70.4 & 56.7 & 75.4  \\
    &  CrossTune & 85.0 & 86.6 & 90.2 & 82.3 &  66.1 & 71.4 & 58.5 & 77.2   \\
    \bottomrule
    \end{tabular}
}
    \caption{Impact analysis of the augmented data amount on the performance of MLP-Classifier and CorssTune. ``Full" refers to the same amount of data as that presented in Table~\ref{tab:data amount}.}
    \label{tab:abl-data-amount}
\end{table*}

\paragraph{Impact of Augmented Data Amount} We study how the performance of CrossTune varies w.r.t. the amount of augmented data. The results of MLP-Classifier are also included in Table~\ref{tab:abl-data-amount} for comparison. Specifically, we compare the cases when the amount of augmented data for each class is 0, 300, and full respectively. Full amount refers to the same setting shown in Table~\ref{tab:data amount}. 


When the quantity of augmented data is 0, i.e., only the original K-shot data is used, the performance of both CrossTune and MLP-Classifier drastically declines by around 10\% on average. The most pronounced performance drop is evident on TREC and QNLI, which contain test cases with diverse semantic and syntactic variations. This observation highlights the importance of boosting the generalization of feature-based black-box tuning approaches with data augmentation. Notably, even without data augmentation, CrossTune performs comparably or better than the prompt-based black-box tuning methods on most datasets (Table~\ref{tab:exp-results}) while requiring no expensive prompt or verbalizer search process. 


After increasing the number of augmented data to 300 per class, the performance on TREC and QNLI improves drastically. The average performance of both MLP-Classifier and CrossTune becomes almost on par with their respective variants trained on the full data, surpassing all the prompt-based black-box methods like BBTv2 and RLPrompt. This suggests that feature-based black-box tuning methods exhibit high data efficiency. 

Finally, regardless of the amount of augmented data used, CrossTune consistently outperforms MLP-Classifier. This further emphasizes the efficacy of utilizing the rich semantics of label descriptions with a cross-attention network. 

\begin{figure*}[!t]
  \centering
  \includegraphics[width=0.8\linewidth]{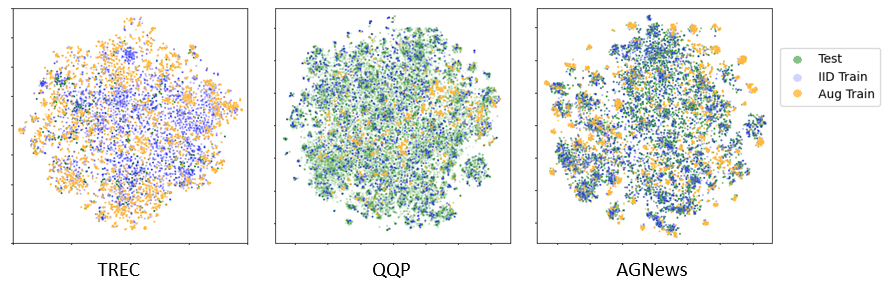}
  \caption{T-SNE Plots of embeddings w.r.t. original training, test, and ChatGPT-augmented training data. Note that we randomly sample the same amount of in-distribution training data as the ChatGPT-augmented data from the original training set.}
  \label{fig:embedding-plot}
\end{figure*}

\begin{figure*}[!t]
  \centering
  \includegraphics[width=0.8\linewidth]{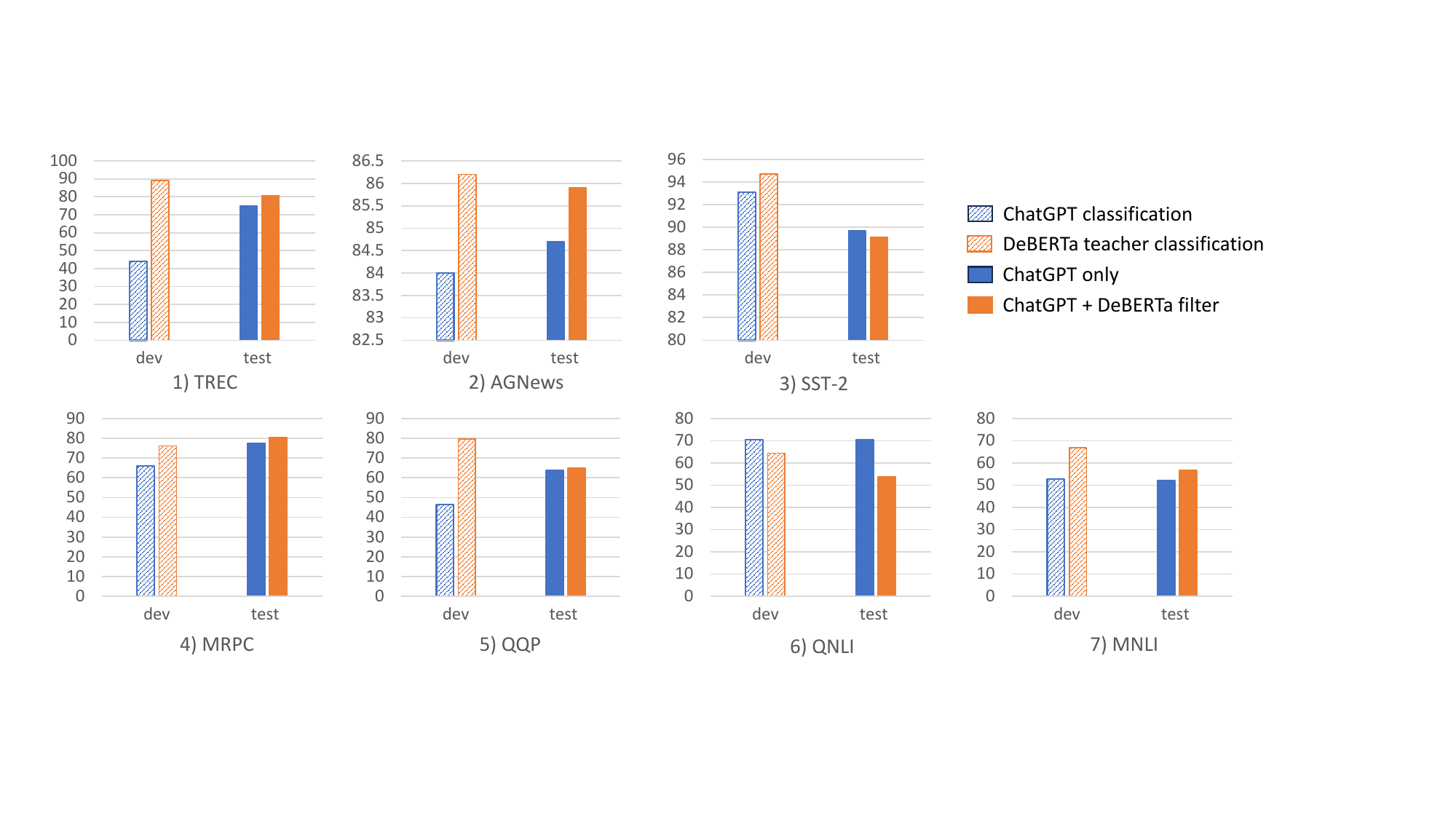}
  \caption{The performance of ChatGPT vs DeBERTa on the development set, which helps determine when to filter the ChatGPT-augmented data. A positive correlation can be observed between the performance of the teacher model on the development set and that of CrossTune on the test set across most datasets.}
  \label{fig:switch-effectiveness}
\end{figure*}

\subsection{Ablation Analysis} 
\label{subsec:ablation-analysis}

\paragraph{In-Distribution vs ChatGPT-Augmented Data} To examine whether ChatGPT is effective in providing augmented data for enhancing the few-shot learners, we compare the performance of learning with the in-distribution 
augmented data against learning with our augmented data using ChatGPT. The in-distribution augmented data are the original task-specific training data with their ground-truth labels removed and then pseudo-labeled with the DeBERTa teacher model. In the case of using ChatGPT-augmented data, we also apply the same DeBERTa teacher to pseudo-label and filter the augmented data. Note that we keep the maximum amount of filtered data the same for both data sources to ensure a fair comparison. Table~\ref{tab:abl-in-vs-out} showcases the results of the MLP-Classifier trained on the two data sources across six different datasets. Except for QQP, training on ChatGPT-Augmented data yields better or comparable results than when trained on in-distribution augmented data. The observation implies that ChatGPT is capable of producing high-quality task-specific data. In practical scenarios, we often have access to limited labeled data and lack in-distribution training data. In these situations, using ChatGPT for data augmentation is a viable option to improve the performance of few-shot learners.

We further analyze the data distributions of ChatGPT-augmented data, the original training data, and the test data. We first encode the text to high-dimensional embeddings with the SimCSE sentence embedder\footnote{\url{https://huggingface.co/princeton-nlp/}}~\citep{gao-etal-2021-simcse} and then apply the T-SNE transformation. Figure~\ref{fig:embedding-plot} shows the plots of TREC, QQP, and AGNews. We can observe that for TREC and AGnews, ChatGPT-augmented data is distributed relatively evenly across the space of the in-distribution training data and resemble a large portion of the test data. However, for QQP, the distribution of ChatGPT-augmented data does not overlap well with the original training data. Besides, because the amount of test data in QQP is much greater than that of the augmented data ($40400\gg1900$), most of the test data are not covered. The observations are in line with results in Table~\ref{tab:abl-in-vs-out} that MLP-Classifier trained on ChatGPT-augmented data performs on par with that trained on the original training data in TREC and AGNews, but worse in the case of QQP. A possible solution is to
optimize the prompts input to ChatGPT such that more diverse data can be generated. We leave such prompt engineering efforts to future work.

\begin{table}[!t]
\centering
\resizebox{\linewidth}{!}{
    \begin{tabular}{l|cccccc|c}
    \toprule
     \multirow{2}{*}{Data Source} & TREC & AGNews & SST-2 & MRPC &  QQP & QNLI & Avg \\
     & acc & acc & acc &  f1 &  f1 & acc & \\
    \midrule
      I.I.D & 78.4 & 86.1 & 88.5 & 75.3 & 77.8 & 66.6 & 78.8 \\
      ChatGPT & 80.8 & 85.9 & 89.1 & 80.4 & 64.8 &  70.4 & 78.6  \\
    \bottomrule
    \end{tabular}
}
    \caption{Performance of MLP-Classifier trained on in-distribution vs ChatGPT-Augmented Data.}
    \label{tab:abl-in-vs-out}
\end{table}

\paragraph{Effectiveness of the Switch} As introduced in~\S\ref{subsec:swich-mechansim}, a switch mechanism is implemented to determine whether to filter the ChatGPT-augmented data with an additional DeBERTa teacher. As depicted in Figure~\ref{fig:switch-effectiveness}, there is a consistent positive correlation between the teachers' performance on the few-shot development set and the final test performance of CrossTune across all the tasks except for SST-2. That is, when the switch is activated (indicating the DeBERTa teacher outperforms ChatGPT), CrossTune, which is trained on data filtered by the DeBERTa teacher, surpasses its variant trained on the unfiltered augmented data, and vice versa. For example, on TREC, AGNews, MRPC, QQP, and MNLI, the test performance of CrossTune improves with DeBERTa filter (as the DeBERTa teacher exhibits superior performance to ChatGPT on the few-shot development set) while on QNLI, the performance of CrossTune is better without the DeBERTa filter, given that the DeBERTa teacher underperforms compared to ChatGPT. These observations confirm that our proposed switch mechanism is reasonable.

\paragraph{Impact of label descriptions} we further study the effect of using different label descriptions. In Table~\ref{tab:different descriptions}, we compare the performance of using long and informative vs short and non-informative label descriptions. When the non-informative descriptions are employed, CrossTune still works but performs slightly worse than when using the long and informative label descriptions. Hence, we can conclude that informative label descriptions help to improve CrossTune's text classification capability. The details of different label descriptions will be presented in appendix in the final version.

\begin{table}[!t]
\centering
\resizebox{0.8\linewidth}{!}{
    \begin{tabular}{l|cccccc}
    \toprule
     \multirow{2}{*}{description type} & TREC & AGNews  & QNLI  \\
     & acc & acc & acc \\
    \midrule
      Short & 83.6 & 85.3 & 70.9 \\
      informative & 85.0 & 86.6 & 71.4 \\
    \bottomrule
    \end{tabular}
}
    \caption{Effect of using informative vs non-informative descriptions.} 
    \label{tab:different descriptions}
\end{table}
\textcolor{black}{Additionally, we examine whether label descriptions also help improve other approaches. Experiments are conducted on MRPC and SST-2 with the MLP-Classifier baseline. Specifically, the input to the model is the concatenation of $\{desc_1, desc_2, \ldots, desc_c, x_i\}$ where $desc_j$ is the label description of the j-th class and $x_i$ is the text sequence to classify. We notice that the performance of the MLP-classifier drops from 89.1\% to 74.65\% on SST-2 and 80.4\% to 75.92\% on MRPC, suggesting a negative impact of label descriptions on the MLP-classifier.}

\paragraph{Impact of Augmented Data on Other Baselines} \textcolor{black}{We further perform experiments with the baselines, BBTv2 and RLprompt on SST2, with the same ChatGPT-augmented data as that used on CrossTune.  No significant improvement is observed compared to training with the original 16-shot data. BBTv2 achieves 83.8\% vs 82.8\% accuracy while RLPrompt achieves 90.5\% vs 91.0\% accuracy before and after data augmentation respectively. It shows these prompt-optimization-based methods do not utilize the augmented data as effectively as CrossTune.} 

\paragraph{Impact of Other Augmentation Techniques}  \textcolor{black}{Besides ChatGPT augmentation, we explore whether traditional data augmentation techniques, also enhance CrossTune. Experiments are conducted with CrossTune on MRPC and SST2, using data augmented from the EDA techniques~\citep{wei-zou-2019-eda}, including random swap, deletion, and insertion of the input text. Our findings indicate that with 300 EDA-augmented data points, CrossTune's performance matches models trained with ChatGPT-augmented data. However, as we increase the data augmentation to 2000, the performance using EDA augmentation deteriorates compared to using no augmentation at all. This decline could be because a significant volume of EDA-augmented data introduces excessive noise into the language model. The deletion, insertion, and swapping operations risk altering the original sentence's semantic meaning. Compared to EDA, ChatGPT-based augmentation emerges as a more reliable method.}

\section{Conclusion}

\noindent In summary, we propose CrossTune for few-shot text classification under the black-box setting. CrossTune treats the black-box LM as a feature extractor and leverages label descriptions as additional input semantic context. To boost the generalization of CrossTune, we avoid relying on in-distribution unlabeled data, instead utilizing ChatGPT to generate pseudo-labeled training samples. A switch mechanism is implemented to ensure the quality of the generated data. Our extensive empirical assessments across seven benchmark datasets reveal CrossTune's effectiveness in black-box tuning, outperforming existing state-of-the-art by an impressive 5.7\% score on average. Even without data augmentation, CrossTune performs better or comparably than previous methods on most datasets.

\section*{Acknowledgement}

We thank the anonymous reviewers for their insightful comments. This work is supported by the National Natural Science Foundation of China (Grant No. 62271432), Shenzhen Science and Technology Research Fund (Fundamental Research Key Project Grant No. JCYJ20220818103001002), and the Internal Project Fund from Shenzhen Research Institute of Big Data under Grant No. T00120220002.

\nocite{*}
\section*{Bibliographical References}\label{sec:reference}

\bibliographystyle{lrec-coling2024-natbib}


\begin{thebibliography}{49}
\expandafter\ifx\csname natexlab\endcsname\relax\def\natexlab#1{#1}\fi

\bibitem[{Bayer et~al.(2022)Bayer, Kaufhold, and Reuter}]{10.1145/3544558}
Markus Bayer, Marc-Andr\'{e} Kaufhold, and Christian Reuter. 2022.
\newblock \href {https://doi.org/10.1145/3544558} {A survey on data augmentation for text classification}.
\newblock \emph{ACM Comput. Surv.}, 55(7).

\bibitem[{Bowman et~al.(2015)Bowman, Angeli, Potts, and Manning}]{bowman-etal-2015-large}
Samuel~R. Bowman, Gabor Angeli, Christopher Potts, and Christopher~D. Manning. 2015.
\newblock \href {https://doi.org/10.18653/v1/D15-1075} {A large annotated corpus for learning natural language inference}.
\newblock In \emph{Proceedings of the 2015 Conference on Empirical Methods in Natural Language Processing}, pages 632--642, Lisbon, Portugal. Association for Computational Linguistics.

\bibitem[{Brown et~al.(2020{\natexlab{a}})Brown, Mann, Ryder, Subbiah, Kaplan, Dhariwal et~al.}]{brown2020language}
Tom Brown, Benjamin Mann, Nick Ryder, Melanie Subbiah, Jared~D Kaplan, Prafulla Dhariwal, et~al. 2020{\natexlab{a}}.
\newblock \href {https://proceedings.neurips.cc/paper/2020/file/1457c0d6bfcb4967418bfb8ac142f64a-Paper.pdf} {Language models are few-shot learners}.
\newblock In \emph{Advances in Neural Information Processing Systems}, volume~33, pages 1877--1901. Curran Associates, Inc.

\bibitem[{Brown et~al.(2020{\natexlab{b}})}]{brown-etal-2020-language}
Tom Brown et~al. 2020{\natexlab{b}}.
\newblock Language models are few-shot learners.
\newblock In \emph{Advances in Neural Information Processing Systems}, volume~33, pages 1877--1901. Curran Associates, Inc.

\bibitem[{Chen et~al.(2021)Chen, Zhang, Zhang, Lee, Cheng, and Li}]{chen-etal-2021-revisiting}
Yiming Chen, Yan Zhang, Chen Zhang, Grandee Lee, Ran Cheng, and Haizhou Li. 2021.
\newblock \href {https://doi.org/10.18653/v1/2021.emnlp-main.718} {Revisiting self-training for few-shot learning of language model}.
\newblock In \emph{Proceedings of the 2021 Conference on Empirical Methods in Natural Language Processing}, pages 9125--9135, Online and Punta Cana, Dominican Republic. Association for Computational Linguistics.

\bibitem[{Cho et~al.(2023)Cho, Kim, and Lee}]{cho-etal-2023-celda}
Hyunsoo Cho, Youna Kim, and Sang-goo Lee. 2023.
\newblock \href {https://doi.org/10.18653/v1/2023.acl-long.239} {{CELDA}: Leveraging black-box language model as enhanced classifier without labels}.
\newblock In \emph{Proceedings of the 61st Annual Meeting of the Association for Computational Linguistics (Volume 1: Long Papers)}, pages 4364--4379, Toronto, Canada. Association for Computational Linguistics.

\bibitem[{Chowdhery et~al.(2022)Chowdhery, Narang, Devlin, Bosma, Mishra, Roberts et~al.}]{chowdhery2022palm}
Aakanksha Chowdhery, Sharan Narang, Jacob Devlin, Maarten Bosma, Gaurav Mishra, Adam Roberts, et~al. 2022.
\newblock \href {http://arxiv.org/abs/2204.02311} {Pa{LM}: Scaling language modeling with pathways}.

\bibitem[{Clark et~al.(2020)Clark, Luong, Le, and Manning}]{clark2020electra}
Kevin Clark, Minh-Thang Luong, Quoc~V. Le, and Christopher~D. Manning. 2020.
\newblock \href {https://openreview.net/forum?id=r1xMH1BtvB} {{ELECTRA}: Pre-training text encoders as discriminators rather than generators}.
\newblock In \emph{International Conference on Learning Representations}.

\bibitem[{Deng et~al.(2022)Deng, Wang, Hsieh, Wang, Guo, Shu, Song, Xing, and Hu}]{deng-etal-2022-rlprompt}
Mingkai Deng, Jianyu Wang, Cheng-Ping Hsieh, Yihan Wang, Han Guo, Tianmin Shu, Meng Song, Eric Xing, and Zhiting Hu. 2022.
\newblock \href {https://doi.org/10.18653/v1/2022.emnlp-main.222} {{RLP}rompt: Optimizing discrete text prompts with reinforcement learning}.
\newblock In \emph{Proceedings of the 2022 Conference on Empirical Methods in Natural Language Processing}, pages 3369--3391, Abu Dhabi, United Arab Emirates. Association for Computational Linguistics.

\bibitem[{Devlin et~al.(2019)Devlin, Chang, Lee, and Toutanova}]{devlin-etal-2019-bert}
Jacob Devlin, Ming-Wei Chang, Kenton Lee, and Kristina Toutanova. 2019.
\newblock \href {https://doi.org/10.18653/v1/N19-1423} {{BERT}: Pre-training of deep bidirectional transformers for language understanding}.
\newblock In \emph{Proceedings of the 2019 Conference of the North {A}merican Chapter of the Association for Computational Linguistics: Human Language Technologies, Volume 1 (Long and Short Papers)}, pages 4171--4186, Minneapolis, Minnesota. Association for Computational Linguistics.

\bibitem[{Diao et~al.(2023)Diao, Huang, Xu, Li, Yong, Zhou, and Zhang}]{diao2023blackbox}
Shizhe Diao, Zhichao Huang, Ruijia Xu, Xuechun Li, LIN Yong, Xiao Zhou, and Tong Zhang. 2023.
\newblock \href {https://openreview.net/forum?id=IvsGP7xRvm} {Black-box prompt learning for pre-trained language models}.
\newblock \emph{Transactions on Machine Learning Research}.

\bibitem[{Du et~al.(2021)Du, Grave, Gunel, Chaudhary, Celebi, Auli, Stoyanov, and Conneau}]{du-etal-2021-self}
Jingfei Du, Edouard Grave, Beliz Gunel, Vishrav Chaudhary, Onur Celebi, Michael Auli, Veselin Stoyanov, and Alexis Conneau. 2021.
\newblock \href {https://doi.org/10.18653/v1/2021.naacl-main.426} {Self-training improves pre-training for natural language understanding}.
\newblock In \emph{Proceedings of the 2021 Conference of the North American Chapter of the Association for Computational Linguistics: Human Language Technologies}, pages 5408--5418, Online. Association for Computational Linguistics.

\bibitem[{Ethayarajh(2019)}]{ethayarajh-2019-contextual}
Kawin Ethayarajh. 2019.
\newblock \href {https://doi.org/10.18653/v1/D19-1006} {How contextual are contextualized word representations? {C}omparing the geometry of {BERT}, {ELM}o, and {GPT}-2 embeddings}.
\newblock In \emph{Proceedings of the 2019 Conference on Empirical Methods in Natural Language Processing and the 9th International Joint Conference on Natural Language Processing (EMNLP-IJCNLP)}, pages 55--65, Hong Kong, China. Association for Computational Linguistics.

\bibitem[{Fei et~al.(2022)Fei, Meng, Nie, Wattenhofer, and Sachan}]{fei-etal-2022-beyond}
Yu~Fei, Zhao Meng, Ping Nie, Roger Wattenhofer, and Mrinmaya Sachan. 2022.
\newblock \href {https://doi.org/10.18653/v1/2022.emnlp-main.587} {Beyond prompting: Making pre-trained language models better zero-shot learners by clustering representations}.
\newblock In \emph{Proceedings of the 2022 Conference on Empirical Methods in Natural Language Processing}, pages 8560--8579, Abu Dhabi, United Arab Emirates. Association for Computational Linguistics.

\bibitem[{Gao et~al.(2021{\natexlab{a}})Gao, Fisch, and Chen}]{gao-etal-2021-making}
Tianyu Gao, Adam Fisch, and Danqi Chen. 2021{\natexlab{a}}.
\newblock \href {https://doi.org/10.18653/v1/2021.acl-long.295} {Making pre-trained language models better few-shot learners}.
\newblock In \emph{Proceedings of the 59th Annual Meeting of the Association for Computational Linguistics and the 11th International Joint Conference on Natural Language Processing (Volume 1: Long Papers)}, pages 3816--3830, Online. Association for Computational Linguistics.

\bibitem[{Gao et~al.(2021{\natexlab{b}})Gao, Yao, and Chen}]{gao-etal-2021-simcse}
Tianyu Gao, Xingcheng Yao, and Danqi Chen. 2021{\natexlab{b}}.
\newblock \href {https://doi.org/10.18653/v1/2021.emnlp-main.552} {{S}im{CSE}: Simple contrastive learning of sentence embeddings}.
\newblock In \emph{Proceedings of the 2021 Conference on Empirical Methods in Natural Language Processing}, pages 6894--6910, Online and Punta Cana, Dominican Republic. Association for Computational Linguistics.

\bibitem[{Hansen and Ostermeier(2001)}]{hansen-ostermeier-2001-completely}
Nikolaus Hansen and Andreas Ostermeier. 2001.
\newblock \href {https://doi.org/10.1162/106365601750190398} {{Completely Derandomized Self-Adaptation in Evolution Strategies}}.
\newblock \emph{Evolutionary Computation}, 9(2):159--195.

\bibitem[{He et~al.(2021)He, Liu, Gao, and Chen}]{he2021deberta}
Pengcheng He, Xiaodong Liu, Jianfeng Gao, and Weizhu Chen. 2021.
\newblock {DEBERTA}: Decoding-enhanced {BERT} with disentangled attention.
\newblock In \emph{International Conference on Learning Representations}.

\bibitem[{Hosseini et~al.(2023)Hosseini, Munia, and Khan}]{hosseini-etal-2023-bert}
MohammadSaleh Hosseini, Munawara Munia, and Latifur Khan. 2023.
\newblock \href {https://doi.org/10.18653/v1/2023.findings-emnlp.1030} {{BERT} has more to offer: {BERT} layers combination yields better sentence embeddings}.
\newblock In \emph{Findings of the Association for Computational Linguistics: EMNLP 2023}, pages 15419--15431, Singapore. Association for Computational Linguistics.

\bibitem[{Hou et~al.(2023)Hou, O'Connor, Andreas, Chang, and Zhang}]{pmlr-v202-hou23b}
Bairu Hou, Joe O'Connor, Jacob Andreas, Shiyu Chang, and Yang Zhang. 2023.
\newblock \href {https://proceedings.mlr.press/v202/hou23b.html} {{P}rompt{B}oosting: Black-box text classification with ten forward passes}.
\newblock In \emph{Proceedings of the 40th International Conference on Machine Learning}, volume 202 of \emph{Proceedings of Machine Learning Research}, pages 13309--13324. PMLR.

\bibitem[{Houlsby et~al.(2019)Houlsby, Giurgiu, Jastrzebski, Morrone, De~Laroussilhe, Gesmundo, Attariyan, and Gelly}]{houlsby-etal-2019-parameter}
Neil Houlsby, Andrei Giurgiu, Stanislaw Jastrzebski, Bruna Morrone, Quentin De~Laroussilhe, Andrea Gesmundo, Mona Attariyan, and Sylvain Gelly. 2019.
\newblock Parameter-efficient transfer learning for {NLP}.
\newblock In \emph{Proceedings of the 36th International Conference on Machine Learning}, volume~97 of \emph{Proceedings of Machine Learning Research}, pages 2790--2799. PMLR.

\bibitem[{Hovy et~al.(2001)Hovy, Gerber, Hermjakob, Lin, and Ravichandran}]{hovy-etal-2001-toward}
Eduard Hovy, Laurie Gerber, Ulf Hermjakob, Chin-Yew Lin, and Deepak Ravichandran. 2001.
\newblock \href {https://aclanthology.org/H01-1069} {Toward semantics-based answer pinpointing}.
\newblock In \emph{Proceedings of the First International Conference on Human Language Technology Research}.

\bibitem[{Hu et~al.(2022)Hu, yelong shen, Wallis, Allen-Zhu, Li, Wang, Wang, and Chen}]{hu2022lora}
Edward~J Hu, yelong shen, Phillip Wallis, Zeyuan Allen-Zhu, Yuanzhi Li, Shean Wang, Lu~Wang, and Weizhu Chen. 2022.
\newblock \href {https://openreview.net/forum?id=nZeVKeeFYf9} {Lo{RA}: Low-rank adaptation of large language models}.
\newblock In \emph{International Conference on Learning Representations}.

\bibitem[{Kahn et~al.(2020)Kahn, Lee, and Hannun}]{jacob-etal-2020-self}
Jacob Kahn, Ann Lee, and Awni Hannun. 2020.
\newblock \href {https://doi.org/10.1109/ICASSP40776.2020.9054295} {Self-training for end-to-end speech recognition}.
\newblock In \emph{ICASSP 2020 - 2020 IEEE International Conference on Acoustics, Speech and Signal Processing (ICASSP)}, pages 7084--7088.

\bibitem[{Lester et~al.(2021)Lester, Al-Rfou, and Constant}]{lester-etal-2021-power}
Brian Lester, Rami Al-Rfou, and Noah Constant. 2021.
\newblock \href {https://doi.org/10.18653/v1/2021.emnlp-main.243} {The power of scale for parameter-efficient prompt tuning}.
\newblock In \emph{Proceedings of the 2021 Conference on Empirical Methods in Natural Language Processing}, pages 3045--3059, Online and Punta Cana, Dominican Republic. Association for Computational Linguistics.

\bibitem[{Li et~al.(2020)Li, Zhou, He, Wang, Yang, and Li}]{li-etal-2020-sentence}
Bohan Li, Hao Zhou, Junxian He, Mingxuan Wang, Yiming Yang, and Lei Li. 2020.
\newblock \href {https://doi.org/10.18653/v1/2020.emnlp-main.733} {On the sentence embeddings from pre-trained language models}.
\newblock In \emph{Proceedings of the 2020 Conference on Empirical Methods in Natural Language Processing (EMNLP)}, pages 9119--9130, Online. Association for Computational Linguistics.

\bibitem[{Li and Liang(2021)}]{li-liang-2021-prefix}
Xiang~Lisa Li and Percy Liang. 2021.
\newblock \href {https://doi.org/10.18653/v1/2021.acl-long.353} {Prefix-tuning: Optimizing continuous prompts for generation}.
\newblock In \emph{Proceedings of the 59th Annual Meeting of the Association for Computational Linguistics and the 11th International Joint Conference on Natural Language Processing (Volume 1: Long Papers)}, pages 4582--4597, Online. Association for Computational Linguistics.

\bibitem[{Liu et~al.(2021)Liu, Zheng, Du, Ding, Qian, Yang, and Tang}]{liu2021gpt}
Xiao Liu, Yanan Zheng, Zhengxiao Du, Ming Ding, Yujie Qian, Zhilin Yang, and Jie Tang. 2021.
\newblock \href {http://arxiv.org/abs/2103.10385} {{GPT} understands, too}.

\bibitem[{Liu et~al.(2019)}]{liu2019roberta}
Yinhan Liu et~al. 2019.
\newblock Ro{BERT}a: A robustly optimized {BERT} pretraining approach.
\newblock \emph{arXiv preprint arXiv: Arxiv-1907.11692}.

\bibitem[{Meng et~al.(2020)Meng, Zhang, Huang, Xiong, Ji, Zhang, and Han}]{meng-etal-2020-text}
Yu~Meng, Yunyi Zhang, Jiaxin Huang, Chenyan Xiong, Heng Ji, Chao Zhang, and Jiawei Han. 2020.
\newblock \href {https://doi.org/10.18653/v1/2020.emnlp-main.724} {Text classification using label names only: A language model self-training approach}.
\newblock In \emph{Proceedings of the 2020 Conference on Empirical Methods in Natural Language Processing (EMNLP)}, pages 9006--9017, Online. Association for Computational Linguistics.

\bibitem[{Mukherjee et~al.(2023)Mukherjee, Mitra, Jawahar, Agarwal, Palangi, and Awadallah}]{mukherjee2023orca}
Subhabrata Mukherjee, Arindam Mitra, Ganesh Jawahar, Sahaj Agarwal, Hamid Palangi, and Ahmed Awadallah. 2023.
\newblock Orca: Progressive learning from complex explanation traces of {GPT-4}.
\newblock \emph{arXiv preprint arXiv: 2306.02707}.

\bibitem[{OpenAI(2023)}]{openai2023gpt4}
OpenAI. 2023.
\newblock \href {http://arxiv.org/abs/2303.08774} {{GPT}-4 technical report}.

\bibitem[{Ouyang and et~al.(2022)}]{ouyang2022training}
Long Ouyang and et~al. 2022.
\newblock Training language models to follow instructions with human feedback.
\newblock In \emph{Advances in neural information processing systems}.

\bibitem[{Prasad et~al.(2023)Prasad, Hase, Zhou, and Bansal}]{prasad-etal-2023-grips}
Archiki Prasad, Peter Hase, Xiang Zhou, and Mohit Bansal. 2023.
\newblock \href {https://doi.org/10.18653/v1/2023.eacl-main.277} {{G}r{IPS}: Gradient-free, edit-based instruction search for prompting large language models}.
\newblock In \emph{Proceedings of the 17th Conference of the European Chapter of the Association for Computational Linguistics}, pages 3845--3864, Dubrovnik, Croatia. Association for Computational Linguistics.

\bibitem[{Raffel et~al.(2020)}]{raffel-etal-2020-exploring}
Colin Raffel et~al. 2020.
\newblock Exploring the limits of transfer learning with a unified text-to-text transformer.
\newblock \emph{Journal of Machine Learning Research}, 21(140):1--67.

\bibitem[{Schick and Sch{\"u}tze(2021)}]{schick-schutze-2021-exploiting}
Timo Schick and Hinrich Sch{\"u}tze. 2021.
\newblock \href {https://doi.org/10.18653/v1/2021.eacl-main.20} {Exploiting cloze-questions for few-shot text classification and natural language inference}.
\newblock In \emph{Proceedings of the 16th Conference of the European Chapter of the Association for Computational Linguistics: Main Volume}, pages 255--269, Online. Association for Computational Linguistics.

\bibitem[{Sohn et~al.(2020)Sohn, Berthelot, Carlini, Zhang, Zhang, Raffel, Cubuk, Kurakin, and Li}]{sohn2020fixmatch}
Kihyuk Sohn, David Berthelot, Nicholas Carlini, Zizhao Zhang, Han Zhang, Colin~A Raffel, Ekin~Dogus Cubuk, Alexey Kurakin, and Chun-Liang Li. 2020.
\newblock Fixmatch: Simplifying semi-supervised learning with consistency and confidence.
\newblock \emph{Advances in neural information processing systems}, 33:596--608.

\bibitem[{Sun et~al.(2022{\natexlab{a}})Sun, He, Qian, Zhou, Huang, and Qiu}]{sun-etal-2022-bbtv2}
Tianxiang Sun, Zhengfu He, Hong Qian, Yunhua Zhou, Xuanjing Huang, and Xipeng Qiu. 2022{\natexlab{a}}.
\newblock \href {https://doi.org/10.18653/v1/2022.emnlp-main.259} {{BBT}v2: Towards a gradient-free future with large language models}.
\newblock In \emph{Proceedings of the 2022 Conference on Empirical Methods in Natural Language Processing}, pages 3916--3930, Abu Dhabi, United Arab Emirates. Association for Computational Linguistics.

\bibitem[{Sun et~al.(2022{\natexlab{b}})Sun, Shao, Qian, Huang, and Qiu}]{sun-etal-2022-bbt}
Tianxiang Sun, Yunfan Shao, Hong Qian, Xuanjing Huang, and Xipeng Qiu. 2022{\natexlab{b}}.
\newblock Black-box tuning for language-model-as-a-service.
\newblock In \emph{Proceedings of the 39th International Conference on Machine Learning}, volume 162 of \emph{Proceedings of Machine Learning Research}, pages 20841--20855. PMLR.

\bibitem[{Wang et~al.(2019)Wang, Singh, Michael, Hill, Levy, and Bowman}]{wang2018glue}
Alex Wang, Amanpreet Singh, Julian Michael, Felix Hill, Omer Levy, and Samuel~R. Bowman. 2019.
\newblock {GLUE}: A multi-task benchmark and analysis platform for natural language understanding.
\newblock In \emph{International Conference on Learning Representations}.

\bibitem[{Wang et~al.(2023)Wang, Kordi, Mishra, Liu, Smith, Khashabi, and Hajishirzi}]{wang-etal-2023-self-instruct}
Yizhong Wang, Yeganeh Kordi, Swaroop Mishra, Alisa Liu, Noah~A. Smith, Daniel Khashabi, and Hannaneh Hajishirzi. 2023.
\newblock \href {https://doi.org/10.18653/v1/2023.acl-long.754} {Self-instruct: Aligning language models with self-generated instructions}.
\newblock In \emph{Proceedings of the 61st Annual Meeting of the Association for Computational Linguistics (Volume 1: Long Papers)}, pages 13484--13508, Toronto, Canada. Association for Computational Linguistics.

\bibitem[{Wei and Zou(2019)}]{wei-zou-2019-eda}
Jason Wei and Kai Zou. 2019.
\newblock \href {https://doi.org/10.18653/v1/D19-1670} {{EDA}: Easy data augmentation techniques for boosting performance on text classification tasks}.
\newblock In \emph{Proceedings of the 2019 Conference on Empirical Methods in Natural Language Processing and the 9th International Joint Conference on Natural Language Processing (EMNLP-IJCNLP)}, pages 6382--6388, Hong Kong, China. Association for Computational Linguistics.

\bibitem[{Xie et~al.(2020)Xie, Dai, Hovy, Luong, and Le}]{xie-etal-2020-uda}
Qizhe Xie, Zihang Dai, Eduard Hovy, Thang Luong, and Quoc Le. 2020.
\newblock \href {https://proceedings.neurips.cc/paper_files/paper/2020/file/44feb0096faa8326192570788b38c1d1-Paper.pdf} {Unsupervised data augmentation for consistency training}.
\newblock In \emph{Advances in Neural Information Processing Systems}, volume~33, pages 6256--6268. Curran Associates, Inc.

\bibitem[{Xu et~al.(2023)Xu, Sun, Zheng, Geng, Zhao, Feng, Tao, and Jiang}]{xu2023wizardlm}
Can Xu, Qingfeng Sun, Kai Zheng, Xiubo Geng, Pu~Zhao, Jiazhan Feng, Chongyang Tao, and Daxin Jiang. 2023.
\newblock Wizard{LM}: Empowering large language models to follow complex instructions.
\newblock \emph{arXiv preprint arXiv: 2304.12244}.

\bibitem[{Zhang et~al.(2021)Zhang, Ding, Xu, Liu, and Zhou}]{zhang-etal-2021-weakly}
Lu~Zhang, Jiandong Ding, Yi~Xu, Yingyao Liu, and Shuigeng Zhou. 2021.
\newblock \href {https://doi.org/10.18653/v1/2021.emnlp-main.222} {Weakly-supervised text classification based on keyword graph}.
\newblock In \emph{Proceedings of the 2021 Conference on Empirical Methods in Natural Language Processing}, pages 2803--2813, Online and Punta Cana, Dominican Republic. Association for Computational Linguistics.

\bibitem[{Zhang et~al.(2023)Zhang, Wang, Zhou, Schuurmans, and Gonzalez}]{zhang2023tempera}
Tianjun Zhang, Xuezhi Wang, Denny Zhou, Dale Schuurmans, and Joseph~E. Gonzalez. 2023.
\newblock \href {https://openreview.net/forum?id=gSHyqBijPFO} {{TEMPERA}: Test-time prompt editing via reinforcement learning}.
\newblock In \emph{The Eleventh International Conference on Learning Representations}.

\bibitem[{Zhang et~al.(2015)Zhang, Zhao, and LeCun}]{zhang-etal-2015-character}
Xiang Zhang, Junbo Zhao, and Yann LeCun. 2015.
\newblock Character-level convolutional networks for text classification.
\newblock In \emph{Advances in Neural Information Processing Systems}, volume~28. Curran Associates, Inc.

\bibitem[{Zhao et~al.(2021)Zhao, Wallace, Feng, Klein, and Singh}]{zhao-etal-2021-calibrate}
Zihao Zhao, Eric Wallace, Shi Feng, Dan Klein, and Sameer Singh. 2021.
\newblock Calibrate before use: Improving few-shot performance of language models.
\newblock In \emph{Proceedings of the 38th International Conference on Machine Learning}, volume 139 of \emph{Proceedings of Machine Learning Research}, pages 12697--12706. PMLR.

\bibitem[{Zoph et~al.(2020)Zoph, Ghiasi, Lin, Cui, Liu, Cubuk, and Le}]{zoph2020rethinking}
Barret Zoph, Golnaz Ghiasi, Tsung-Yi Lin, Yin Cui, Hanxiao Liu, Ekin~Dogus Cubuk, and Quoc Le. 2020.
\newblock Rethinking pre-training and self-training.
\newblock \emph{Advances in neural information processing systems}, 33:3833--3845.

\end{thebibliography}


\appendix

\section{Addtional Dataset Details}
\label{sec:additional-details}

\begin{table}[!h]
\centering
\resizebox{\linewidth}{!}{
    \begin{tabular}{l|ll}
    \toprule
    \textbf{Dataset} & \textbf{Label} & \textbf{Description} \\ \midrule
     \multirow{6}{*}{TREC} & description & Answer to the question is a description. \\
     & entity & Answer to the question is an entity. \\
     & abbreviation & Answer to the question is an abbreviation. \\
     & number & Answer to the question is a number. \\
     & human & Answer to the question is a human. \\
     & location & Answer to the question is a location. \\
    \midrule
    \multirow{4}{*}{AGNews} & tech & It is a technology news.   \\
     & world & It is a world news.  \\
     & sports & It is a sports news.  \\
     & business & It is a business news.  \\
    \midrule
    \multirow{2}{*}{QNLI} & entailment & The statement contains the answer to the question.  \\
     & non\_entailment & The statement contains no answer to the question.  \\
    \bottomrule
    \end{tabular}
}
    \caption{Short and Non-informative Label Descriptions} 
    \label{tab:non-informative-description}
\end{table}

Table~\ref{tab:non-informative-description} depicts the short and non-informative label descriptions used in our ablation study (\S\ref{subsec:ablation-analysis}) where we compare the effects of using informative label descriptions against using non-informative ones. Table~\ref{tab:informative-description} shows the label descriptions we use in our main experiments. 

\begin{table*}[!ht]
\centering
\resizebox{\linewidth}{!}{
    \begin{tabular}{c|c|l}
    \toprule
    \textbf{Dataset} & \textbf{Label} & \textbf{Description} \\ \midrule
     \multirow{6}{*}{TREC} & description & \thead[l]{Definition of something, description of something, manner of an action, reason.} \\ 
     & entity & \thead[l]{Animal, Organ of body, Color, Invention, book and other creative piece, Currency name, Disease and medicine, \\ Event, Food,  Musical instrument, Language, Letter like a-z, Other entity, Plant, Product, Religion, Sport, Element \\ and substance, Symbols and sign, Techniques and method, Equivalent term, Vehicle, Word with a special property.} \\
     & abbreviation & \thead[l]{A shortened form of a word or phrase that is used to represent the full meaning.}  \\ 
     & number & \thead[l]{Number of something, Date, Distance, Price, Order, rank, Lasting time, Percent, \\ fraction, Speed, Temperature, Size, area and volume, Weight, Postcode or other code.}  \\
     & human & \thead[l]{Individual, Title of a person, Description of a person, Group or organization of persons.}  \\ 
     & location & City, Country, Mountain, State, Other location.   \\
    \midrule
    \multirow{4}{*}{AGNews} & tech & \thead[l]{The Sci/Tech category is designed to encompass articles related to science and technology. It might include news about \\ scientific discoveries or research breakthroughs, technology product launches, technology company updates, coverage of \\ scientific and technology conferences, interviews with scientists or tech leaders, articles on new theories or models \\ in various scientific disciplines, advancements in medical technology, and many more.}   \\
     & world & \thead[l]{It's a news article about international affairs, geopolitics, global events, or any topic that has a worldwide or international scope. \\ Examples may include news on international diplomacy, major global events like the United Nations General Assembly, \\ international conflicts  or wars, significant elections or political events in different countries, global environmental issues, and more. }   \\
     & sports & \thead[l]{Articles related to various sporting events, news, and updates. the Sports category could encompass \\ a wide range of topics such as game results, player transfers, injuries, interviews with athletes, coverage of international \\ sporting events like the Olympics, football (soccer) world cup, tennis grand slams, and more.} \\
     & business & \thead[l]{The Business category typically cover topics related to commerce, economics, and finance on a local, national, or international \\ scale. It may include news about company mergers, financial reports, stock market updates, changes in economic  \\ policies, interviews with business leaders, innovation in business models, trends in various industry sectors, and so on.}  \\
    \midrule
    \multirow{2}{*}{QNLI} & entailment & \thead[l]{The given statement logically contains the answer to the associated question. \\ If the truth of the statement provides the answer to the question, it's considered an entailment.}\\
     & non\_entaiment & \thead[l]{The given statement does not logically contain the answer to the associated question. \\ Even if the statement is true, it does not provide a valid answer to the question.}   \\  \midrule
     \multirow{3}{*}{MNLI} & entailment & \thead[l]{The hypothesis can be logically inferred or implied from the premise.} \\
        & neutral & \thead[l]{The premise and the hypothesis do not have a clear logical relationship.} \\
        & contradiction & \thead[l]{The hypothesis contradicts or conflicts with the information presented in the premise.}  \\ \midrule
    \multirow{2}{*}{MRPC} & equivalent & \thead[l]{Two sentences in the pair are semantically equivalent - they express the same, or very similar, meaning.}  \\
     & non\_equivalent & \thead[l]{Two sentences in the pair are not semantically equivalent - they do not convey the same meaning.} \\ \midrule
    \multirow{2}{*}{QQP} & equivalent & \thead[l]{That's to say,}   \\
     & non\_equivalent & \thead[l]{Another different questions is,} \\  \midrule
    \multirow{2}{*}{SST-2} & positive & \thead[l]{sentences from movie reviews that express favorable, complimentary, or praiseworthy viewpoints about a movie. \\ The concept of positive sentiment in this context typically includes feelings of enjoyment, admiration, appreciation, \\ or satisfaction with elements of a movie such as its plot, acting, direction, cinematography, or other aspects of its production.}   \\
     & negative & \thead[l]{Sentences that express unfavorable, critical, or disparaging viewpoints about a movie. The concept of negative \\ sentiment here typically includes feelings of disappointment, dissatisfaction, frustration, or displeasure with elements \\ of a movie such as its plot, acting, direction, cinematography, or other aspects of its production.} \\
    \bottomrule
    \end{tabular}
}
    \caption{Label Descriptions used in main experiments} 
    \label{tab:informative-description}
\end{table*}

\end{document}